\documentclass[letterpaper]{article} 
\usepackage{aaai19}  
\usepackage{times}  
\usepackage{helvet}  
\usepackage{courier}  
\usepackage{url}  
\usepackage{graphicx}  
\usepackage{microtype}
\usepackage{booktabs}
\usepackage{amssymb}

\newcommand{\lp}{\left}
\newcommand{\rp}{\right}

\nocopyright

\usepackage{amsmath} 
\usepackage{color}
\usepackage{caption}
\usepackage{subcaption}
\usepackage{multirow}
\newcommand{\citeay}[1]{\citeauthor{#1} (\citeyear{#1})}

\title{Generalized Batch Normalization: Towards Accelerating Deep Neural Networks}
\author{Xiaoyong Yuan\thanks{Equal Contribution} \\ University of Florida \\ chbrian@ufl.edu \And Zheng Feng\footnotemark[1] \\University of Florida \\ fengzheng@ufl.edu \And Matthew Norton \\ Naval Postgraduate School \\ mnorton@nps.edu \And Xiaolin Li \\ University of Florida \\ andyli@ece.ufl.edu}

\begin{document}

\maketitle
\begin{abstract}
Utilizing recently introduced concepts from statistics and quantitative risk management, we present a general variant of Batch Normalization (BN) that offers accelerated convergence of Neural Network training compared to conventional BN. In general, we show that mean and standard deviation are not always the most appropriate choice for the centering and scaling procedure within the BN transformation, particularly if ReLU follows the normalization step. We present a Generalized Batch Normalization (GBN) transformation, which can utilize a variety of alternative \textit{deviation measures} for scaling and \textit{statistics} for centering, choices which naturally arise from the theory of \textit{generalized deviation measures} and risk theory in general. When used in conjunction with the ReLU non-linearity, the underlying risk theory suggests natural, arguably optimal choices for the deviation measure and statistic. Utilizing the suggested deviation measure and statistic, we show experimentally that training is accelerated more so than with conventional BN, often with improved error rate as well. Overall, we propose a more flexible BN transformation supported by a complimentary theoretical framework that can potentially guide design choices.
\end{abstract}


\section{Introduction}
Training a deep neural network has traditionally been a difficult task. Issues such as the vanishing and exploding gradient, see e.g., \cite{pascanu2013difficulty}, make the use of gradient based optimization techniques difficult from the perspective of stability and fast convergence. However, new, seemingly simple tools have emerged to help practitioners overcome common pitfalls of neural network training. Two prominent examples are the use of Batch Normalization (BN) and Rectified Linear Units (ReLU).

Originally proposed by \cite{ioffe2015batch}, BN provides a simple transformation which incentivizes the homogenization of neural network layer outputs, so as to have the same scale and mean, eliminating what is referred to as internal covariate shift. Intuitively, this allows the `signal' flowing through the neural network to maintain a consistent center and scale, potentially stabilizing gradients and the training procedure as a whole.

Consider a single layer of the network which first receives output from the previous layer $h$ and then applies an affine transformation to get $\mathbf x = Wh + b$, followed by an element-wise non-linearity to produce output $g(\mathbf x)$ which is fed to the next layer. Let $\mathbf x=[ x_1, x_2,\cdots,x_{n}]^T$ denote its individual components so that we can write $g(\mathbf x)=[ g(x_1), g(x_2),...,g(x_{n})]^T$. 

The BN transformation is based upon the following transformation on each dimension $j$ of the input,
$$\hat{x}_j \leftarrow \frac{x_j - E[x_j]}{\sqrt{E[ (x_j-E[x_j])^2 ]}}\;,$$
where $E[x_j]$ and $\sqrt{E[ (x_j-E[x_j])^2 ]}$ are the mean and standard deviation of the random variable $x_j$, which are estimated during training with a batch of training examples. 

In this paper, we begin by asking the question: Are mean and standard deviation the right choice for every network architecture? This simple question leads us to the main contribution of this paper, which is the observation that batch normalization can naturally be generalized and improved by considering the general transformation,
$$\hat{x}_j\leftarrow \frac{x_j - \mathcal{S}(x_j)}{\mathcal{D}(x_j)} \;,$$
where $\mathcal{D}$ is some measure of \textit{deviation}, not necessarily the standard deviation, and where $\mathcal{S}$ is a \textit{statistic} which is not necessarily the mean. While arising from a specific set of axioms in risk theory, one can think of $\mathcal{D}$ as a general measure of the non-constancy of $x$ and $\mathcal{S}$ as a type of 'center.`

We show that there exist many different choices for $\mathcal{D}$ and $\mathcal{S}$ besides standard deviation and mean, and that by formulating the batch normalization transformation with these alternatives one can accelerate neural network training compared to conventional BN and, in some settings, obtain improved predictive performance. Additionally, we show how the choice of $\mathcal{D}$ and $\mathcal{S}$ are driven not only by straightforward intuition, but also by recently developed theoretical tools from statistics and risk theory. Specifically, the theory of \textit{generalized deviation measures} provides us with a wealth of choices for \textit{deviation measure} $\mathcal{D}$, which includes standard deviation as a special case. In addition, for any choice of $\mathcal{D}$, there is a naturally corresponding \textit{statistic} $\mathcal{S}$. Thus, choosing $\mathcal{D}$ implies natural choices for $\mathcal{S}$ and vice versa. 

Besides the simple observation that mean and standard deviation can be replaced by alternatives, our analysis is also driven by the observation that the appropriateness of the choice of $\mathcal{D}$ and $\mathcal{S}$ is directly tied to the choice of non-linearity which follows the normalization transformation. We focus our analysis on the ReLU non-linearity from \cite{glorot2011deep} and \cite{nair2010rectified}, which has played a significant role in stabilizing and accelerating neural network training~\cite{krizhevsky2012imagenet,dahl2013improving}. We show that mean and standard deviation are not natural choices for centering and scaling if ReLU follows the normalization transformation. Risk theory and simple intuition suggest more natural choices. In fact, we see that one of these choices, the superquantile deviation, allows explicit control over the level of sparsity of activation's; hypothesized to be an important property of ReLU \cite{glorot2011deep}. While we focus on ReLU, this intuition can also be applied to any asymmetric non-linearity such as Leaky ReLU \cite{maas2013rectifier}, Exponential Linear Unit \cite{clevert2015fast}, or any other arising from the ReLU family (e.g. \cite{he2015delving}).

We demonstrate on MNIST, CIFAR-10, CIFAR-100, and SVHN datasets that the speed of convergence of stochastic gradient descent (SGD) can be increased by simply choosing a different $\mathcal{D}$ and $\mathcal{S}$ and that, in some settings, we obtain improved predictive performance. Our experimental analysis also serves to support the intuition that ReLU paired with $\mathcal{D}=\sqrt{E[ (x-E[x])^2 ]}$ and $\mathcal{S}=E[x]$ is a mismatch and that asymmetric choices for $\mathcal{D}$ and $\mathcal{S}$ which are suggested by risk theory and intuition do, in fact, work better.

Although much further analysis is needed in this direction, we show that the use of ReLU's in tandem with BN can be tied directly to risk theory via a recently introduced concept called Buffered Probability of Exceedance (bPOE). Specifically, the use of normalization followed by a ReLU gives rise to what can be considered to be the tightest convex approximation to the $0-1$ loss. This is intriguing given the history of neural networks began with the concept of $0-1$ loss (indicator function) neural output which were then approximated with the sigmoid transformation as a differentiable surrogate (see e.g. \cite{rosenblatt1958perceptron,mcculloch1943logical}). 

\section{Batch Normalization}
The BN transformation is based upon the following transformation on each dimension $j$ of the input,
$$\hat{x}_j \leftarrow \frac{x_j - \mu_j}{\sqrt{\sigma_j^2 + \epsilon}}\;,$$
where $\sigma_j$ and $\mu_j$ are the empirical standard deviation and mean of the random variable $x_j$, which are estimated during training with a batch of training examples. Throughout this paper, we will view $\mathbf x$ as a random vector which is observed empirically via the training batches. Thus, during training, $\mu_j = \frac{1}{|B|} \sum_{i=1}^{|B|} x^{(i)}_{j}$ with $|B|$ denoting the size of the training batch.

The BN procedure follows the actual normalization with the following linear transformation, where $\gamma_j,\beta_j$ are parameters which will be tuned during training,
$$ \gamma_j \hat{x}_j + \beta_j \;.$$
The BN procedure is then followed by the final non-linear transformation $g(\gamma_j \hat{x}_j + \beta_j)$. Why is this linear transformation needed? As noted by \cite{ioffe2015batch}, the BN transformation may not be appropriate or work well in conjunction with the non-linear transformation $g$ that follows. Thus, the authors introduced a way to adjust the BN transformation if necessary. However, there is no guarantee that training will find the right linear transformation and be able to properly counteract a poor choice of scale and center. In some sense, this is why it is argued in \cite{mishkin2015all} that proper initialization is all that is needed. Assuming that the centering and scaling are not correct, which is to say that the trainable linear transformation is necessary to adjust the center and scale, then BN can be loosley viewed as a type of data dependent initialization strategy. In this sense, the additional linear transformation can be used within our proposed scheme in exactly the same way, but with more control over the initialization where one would hope to select a more appropriate data-dependent centering and scaling factor.

Cases where the standard BN may not work well in conjunction with the non-linearity $g$ can be easily illustrated, particularly if $g$ is the ReLU non-linearity. Consider a set of outputs $\{x_1,\cdots,x_N\}$ from a network layer which have mean zero, i.e., $\frac{1}{N} \sum_{i=1}^N x_i=0$, with ordering $x_1<\cdots<x_k<0<x_{k+1}<\cdots<x_N$. Assume that we are then going to divide by some normalization factor, such as standard deviation, and then feed these values into a ReLU non-linearity $\max\{0,x_i\}$. The ReLU non-linearity will map points $x_1,\cdots,x_k$ to zero. Considering this fact, does it make sense to first divide the whole set of $N$ points by the standard deviation? Intuitively, it would make more sense to divide by the variance of only the set of points $\{x_{k+1},\cdots,x_N\}$. The variation of the set of points $\{x_1,\cdots,x_k\}$ is irrelevant given the fact that a ReLU will follow, sending all of these points to zero. This consideration is particularly important if the conditional distributions $\{x_1,\cdots,x_k\}$ and $\{x_{k+1},\cdots,x_N\}$ exhibit very different scales and variation. In this case, it may be more appropriate to use a one-sided measure of deviation $\mathcal{D}$ for the normalization step such as the Right Semi-Deviation (RSD) $\frac{1}{N}\sum_{i=1}^N \max\{0,x_i\}$. Furthermore, a similar argument can be applied to the centering operation. Assume, for instance, that the distribution $\{x_1,\cdots,x_N\}$ is heavy-tailed, with $\{x_1,\cdots,x_{N-1}\}$ having mean zero and variance 1, but with $x_N=100$. The mean of all $N$ points will be very large, and centering the data via mean subtraction will yield $x_N-\mu$ as the only term with value larger than zero. Thus, the application of the ReLU will leave only one sample as having non-zero value (and gradient), with much of the valuable learning signal lost because of poor choice of centering statistic $\mathcal{S}$.

This paper shows that there are other ways to perform batch normalization, potentially avoiding the need to adjust the normalization with the affine transformation (or at least reducing the amount by which it would need to be adjusted), offering accelerated convergence. 
Generalizations and variants of BN have been proposed before. For example, \citeay{klambauer2017self} proposed a self-normalizing network layer, but is limited to standard feed-forward architectures. \citeay{ba2016layer} altered BN to work with recurrent neural networks. \citeay{mishkin2015all} argue that BN is simply another way to perform initialization, thus proposing initialization methods that produce similar effects. The idea of BN was altered to weight normalization by reparameterizing the weights~\cite{salimans2016weight,chunjie2017cosine}. Our proposed approach, while relying on simple principles, is grounded in a broader theory and maintains all important flexibility of conventional BN.

\section{Asymmetric Deviation Measures in Risk Theory}
As alluded to in the introduction, it is easy to question the use of variance as the scale normalizing factor if it is followed by the ReLU transformation. This gives rise to the obvious question: What other options do we have that may be more appropriate? We find, in general, that risk theory provides us with an entire class of generalized deviation measures to choose from. In this section, we briefly introduce risk theory before discussing generalized deviation measures in Section 4 where we introduce the GBN transformation and show that generalized deviation measures provide us with an array of alternatives to mean and standard deviation.

Over the past 25 years, risk management theory has played a crucial role in the development of fundamental statistical concepts that not only measure risk \cite{artzner1999coherent,follmer2002convex,szego2002measures}, but have proven fundamental to statistical theory and optimization under uncertainty. A full review of risk theory is beyond the scope of this paper, but a simple example in the context of financial engineering can be used to illustrate. Consider an investment which will yield a loss of $x$, with $x$ being a random monetary loss. Assume we knew the distribution of $x$, and we were to ask: How \textit{risky} is this investment? How can we measure \textit{risk} to compare it against other investments $y$? An obvious choice would be to look at the expected loss $E[x]$. However, this may be inappropriate, as investor objectives (or distribution of $x$) may be highly asymmetric. It may be more appropriate to measure risk with an asymmetric quantity. One example would be to use the quantile $q_{\alpha}(x)=\min \{z | P(x\leq z)\geq \alpha \}$, where $\alpha \in [0,1]$ is a probability level. Its inverse, called Probability of Exceedance (POE), given by $P(x>z)$ where $z \in \mathbb{R}$ is some known threshold, may also be desirable if some threshold $z$ is known and exceeding such a threshold is undesirable. 

One of the primary drivers of risk theory, however, has been the need to quantify risk in such a way that optimization can take place (e.g. finding the portfolio with minimal risk). The quantile, also called the Value-at-Risk, and POE are numerically troublesome in this context. Specifically, these functions often prove to be non-convex and discontinuous, essentially reducing to sums of indicator ($0-1$ loss) functions. From this difficulty, more amenable alternatives have arisen. 

Two popular alternatives that are relevant to our discussion are the superquantile and Buffered Probability of Exceedance (bPOE) \cite{Uryasev,rockafellar2002conditional,acerbi2002coherence,Mafusalov}. The superquantile is a measure of uncertainty similar to the quantile, but with superior mathematical properties. Formally, the superquantile, also called Conditional Value-at-Risk (CVaR) in the financial engineering literature, for a continuously distributed $x$ is defined as
$$\bar{q}_\alpha (x) =E \lp[ x | x > q_\alpha (x) \rp].$$
For general distributions, the superquantile can be defined by the following formula,
\begin{equation}
\label{CVaR}
\bar{q}_\alpha (x)=\min_{\gamma} \gamma + \frac{ E[ x-\gamma ]^+} {1-\alpha} \; ,
\end{equation}
where $[\cdot]^+=\max\{\cdot,0\}$.

Similar to $q_\alpha (x)$, the superquantile can be used to assess the tail of the distribution. The superquantile, though, is far easier to handle in optimization contexts. It also has the important property that it considers the magnitude of events within the tail. Therefore, in situations where a distribution may have a heavy tail, the superquantile accounts for magnitudes of low-probability large-loss tail events while the quantile does not account for this information.

bPOE is the inverse of the superquantile. In other words, bPOE calculates one minus the probability level at which the superquantile equals a specified threshold $z$. It is calculated by the formula
$$ \bar{p}_z (x) = \min_{a \geq 0} E[ a(x-z) +1 ]^+ = \min_{\gamma <z} \frac{ E[x - \gamma]^+ } { z - \gamma},$$
where $[\cdot]^+ = \max\{\cdot , 0\}.$ In addition, we have the following formula which will be important for our case. Assuming that $\bar{p}_z (x) =1-\alpha$, we have that
$$ \bar{p}_z (x) =   \frac{ E[x - q_\alpha(x)]^+ } { \bar{q}_\alpha(x) - q_\alpha(x)} \;.$$
Roughly speaking, bPOE calculates the proportion of worst case outcomes which average to $z$. 

As it relates to POE, bPOE can be viewed as an optimal convex approximation. More specifically, among law-invariant functions of $x$, $ \bar{p}_z(x)$ is the minimal (tightest) quasi-convex upper bound of $P(x>z)=E[I(x>z)]$.


These ideas, though, have not been limited to risk management and finance. Machine Learning has also been impacted by this theory. For example, new support vector classifiers have been generated with superquantile and bPOE concepts \cite{takeda2008nu,norton2017soft,gotoh2017support} and sequential decision problems are being formulated with risk in mind \cite{galichet2013exploration,chow2014algorithms}.

\section{Generalized Batch Normalization}
In this paper, we define Generalized Batch Normalization (GBN) to be identical to conventional BN but with standard deviation replaced by a more general deviation measure $\mathcal{D}(x)$ and the mean replaced by a corresponding statistic $\mathcal{S}(x)$. In other words, we have the transformation,
$$\hat{x}_j\leftarrow \frac{x_j - \mathcal{S}(x_j)}{\mathcal{D}(x_j)} \;.$$
Here, each choice of $\mathcal{D}$ is naturally paired with some $\mathcal{S}$, which we discuss in the following section. In Section 5, we implement a suite of these new measures and test them on the MNIST, CIFAR-10, CIFAR-100, and SVHN datasets, showing that convergence can be accelerated, and sometime accuracy improved, by use of different deviation measures and statistics.
\subsection{Generalized Deviation Measures and Statistics}
\begin{table*}[!tb]
\centering
\small
\begin{tabular}{ l | c | r }
  \hline			
   & Deviation Measure $\mathcal{D}(x)$& Statistic $\mathcal{S}(x) $\\ \hline
  Standard Deviation (SD)& $\sqrt{E[ (x-E[x])^2 ]}$ & $E[x]$ \\
  Mean Absolute Deviation (MAD) & $E[ |x-E[x]| ]$ & $E[x]$ \\
  Right-Semi-Deviation (RSD) & $E[ x-E[x] ]^+$ & $E[x]$ \\
  Superquantile Deviations (SQD) for $\alpha \in (0,1)$ & $\bar{q}_\alpha( x - E[x]) $ & $q_\alpha (x) $\\
  Range-Based Deviation (RBD) &  $  \sup x - \inf x$ & $ \frac{1}{2}(\sup x + \inf x)$ \\
  Worst-Case Deviation (WCD) & $ \sup x - E[x] $ & $ \sup x $ \\
  \hline  
\end{tabular}
\caption{Examples of deviation measures and their corresponding statistics.$[x]^+=\max\{0,x\}$}
\label{risk_measure_table}
\end{table*}
In \cite{rockafellar2006generalized}, the concept of a generalized deviation measure was introduced to broaden the statistical view of \textit{deviation} beyond the single case of standard deviation, specifically for use in quantitative risk analysis. These deviation measures follow a very general set of axioms which we will not delve into here. However, some examples can be found in Table~\ref{risk_measure_table}, and they can be understood intuitively as follows: Deviation measures quantify the \textit{non-constancy} of a random variable. As seen in Table~\ref{risk_measure_table}, standard deviation is only one of many possibilities, such as the asymmetric deviation measures RSD and SQD with $\alpha>0$. These measures of deviation look at the variation only in the \textit{right-tail} of the distribution of $x$. It's easy to see how this type of asymmetric measure would be of interest in finance, where it may be important to analyze the variation of only the largest losses within the \textit{right-tail}.

The theory of generalized deviation measures is also complemented by the recently introduced theory of the Risk Quadrangle. Utilizing functional relationships that are beyond the scope of this paper, \cite{rockafellar2013fundamental} shows that measures of deviation are intimately related to similar measures of risk, regret, and error. Furthermore, associated with any measure of deviation is a unique \textit{statistic}. In short, however, without getting into too much detail, one can think of the statistic as a type of `center.' In Table~\ref{risk_measure_table}, we see how this intuition plays out, with the corresponding statistics listed in the right column. For SD, MAD, and RSD, we see that $\mathcal{S}(x)$ is simply the expectation. However, for SQD with $\alpha=.5$, we see that $\mathcal{S}(x)=q_{.5}(x)$ the median, certainly a different notion of the `center.' Furthermore, we see that for RBD, the statistic is the center of the range. However, for SQD it is important to notice that we can achieve very different statistics by moving $\alpha$, which gives us different quantiles.
\subsection{Choosing $\mathcal{D}$ or $\mathcal{S}$: General Intuition}
Now that we are given more options for deviation measures and statistics, we can begin to think about the benefits and drawbacks of each within the neural network architecture and the GBN transformation. Utilizing standard deviation seems like an intuitive choice. However, this depends heavily on the shape of the (empirical) distribution of $x$. If the distribution is relatively symmetric, then standard deviation will be indicative of the overall scale and the mean will be indicative of the `center'. Similarly, this may hold true if the distribution does not have heavy tails or outliers on one side or the other. However, if the distribution of $x$ has e.g. heavy tails, is highly asymmetric, has outliers, or is multimodal; then the mean may be a poor choice for the `center' and the deviation of values to the right of the mean may be dramatically different than the deviation of values to the left of the mean. In this case, a quantile may be a more appropriate notion of the `center.' Choosing, for example, the median instead of the mean assures that we are truly `centering' the data, with half of the points on the `left' and half on the `right.'

Even if the distribution of $x$ is not asymmetric or heavy tailed, the choice of center is particularly important if normalization is followed by the ReLU activation. Specifically, the choice of center controls the sparsity of activation's produced by the ReLU, since any elements left-of-center will be sent to zero. ReLU induced sparsity has been hypothesized as critical to its success \cite{glorot2011deep}. In this case, the quantile is a natural choice for center that provides precise control over such sparsity. If the normalization centers w.r.t. the quantile at $\alpha$, exactly $\alpha\%$ of activation's across the batch will have zero value.  


Driving our intuition from the beginning was the idea that the non-linearity, deviation measure, and statistic should be chosen in tandem. As mentioned in Section 2, the pairing of ReLU with typical BN (i.e. standard deviation and mean normalization) does not seem appropriate given the fact that standard deviation is symmetric while ReLU is asymmetric. Thus, in light of Section 2, we find that asymmetric deviation measures are more appropriate such as RSD or SQD for any $\alpha >0$. In Section 5, we see this intuition confirmed, with RSD and SQD outperforming SD in terms of convergence rate and, often times, test error. Although not explored in our experiments, this intuition applies to any asymmetric non-linearity such as the Leaky ReLU \cite{maas2013rectifier}, Exponential Linear Unit \cite{clevert2015fast}, or any other arising from the ReLU family (e.g. \cite{he2015delving}).

\subsection{An Optimal Choice}
Beyond this simple intuition, we can utilize connections to risk theory to provide evidence that the ReLU should be used in tandem with an asymmetric deviation measure. Specifically, we show that the use of SQD and RSD followed by ReLU is approximately equivalent to a probabilistic transformation which mimics an optimal quasiconvex approximation to the $0-1$ (indicator) loss function. 


Intuitively, ReLU's should be paired with an asymmetric measure of deviation, with candidates including RSD and SQD. However, a natural choice arises when looking at the similarities between bPOE and the combination of the GBN transformation and ReLU non-linearity.
Consider a GBN transformation followed by a ReLU non-linearity. Now, for the GBN transformation let us choose SQD deviation measure $\mathcal{D}(x_j)=\bar{q}_\alpha (x_j - \mu_j )$ where $\alpha$ is chosen so that $q_\alpha(x_j) = \mu_j$, meaning that we are choosing the probability level on which the mean sits. This gives us the following transformation, where the superscript denotes the $i^{th}$ sample from a batch:
$$\hat{x}^{(i)}_j \leftarrow \lp[ \frac{x^{(i)}_j  - q_\alpha(x_j) }{ \bar{q}_\alpha (x_j ) - q_\alpha(x_j) } \rp]^+\;.$$
This can be re-written as,
$$\hat{x}^{(i)}_j \leftarrow \lp[ \frac{x^{(i)}_j  - \mu_j }{ E[x_j - \mu_j | x_j > \mu_j] }\rp]^+  \;.$$
One will immediately notice that this is almost identical to a conventional BN transformation followed by ReLU with the only difference being that we are dividing by a one-sided semi-deviation rather than the two-sided standard deviation. 
One will notice, however, the following connection to bPOE: 
$$\bar{p}_z (x_j) =  E\lp[ \frac{x_j - q_\alpha(x_j) }{ \bar{q}_\alpha (x_j ) - q_\alpha(x_j) }\rp]^+ $$
for threshold $z= \bar{q}_\alpha (x_j )$. Thus, we see that the combination of GBN and ReLU yields a transformation based upon bPOE. If also divided by sample size $N$, each individual sample $x^{(i)}_j$ will yield output $\frac{1}{N} \lp[ \frac{x^{(i)}_j - q_\alpha(x_j) }{ \bar{q}_\alpha (x_j ) - q_\alpha(x_j) } \rp]^+ \in (0,1)$ with the sum,
\[ 
 \frac{1}{N} \lp[ \frac{x^{(i)}_j - q_\alpha(x_j) }{ \bar{q}_\alpha (x_j ) - q_\alpha(x_j) } \rp]^+ =  \bar{\hat{p}}_z (x_j)   \;,
\]
where $\bar{\hat{p}}_z (x_j)$ simply denotes the empirical bPOE calculated from a sample. This means that the overall output distribution will consist of values in the range $[0,1]$ with non-zero items being those that are in the bPOE-tail of the empirical distribution of $x_j$.

Thus, by combining GBN and ReLU we are effectively performing a probabilistic transformation, with the transformation mimicking the optimal quasiconvex approximation to the $0-1$ loss. 

\section{Experimental Evaluation}
Overall, the first goal of our experiments is to demonstrate the obvious: All other things being equal, different normalization methods (i.e. different choices for deviation measure and statistic) lead to different network properties. We then explore the specifics of these changes. First, we show that convergence rate and stability of NN training via SGD can often be improved by utilizing alternative deviation measures. Improvement is measured relative to conventional BN, which uses mean and standard deviation as its statistic and deviation measure. Overall, we find that SQD, MAD, and RSD often lead to increased convergence rates and, sometimes, increased stability in terms of smoothly decreasing test error during SGD. Second, we see that these alternative choices often lead to testing error that is nearly as good as, or better, than that achieved by standard BN.

For all experiments, GBN is implemented in exactly the same manner as standard BN, only with mean and variance replaced by generalized $\mathcal{S}$ and $\mathcal{D}$ within the batch normalization transformation. This includes appropriate inclusion of the chosen deviation measure and statistic within the gradient calculation as well as the batch-based estimation of $\mathcal{D}(x_j)$ and $\mathcal{S}(x_j)$ during training and population-based estimation for inference. This also includes the additional linear transformation which typically follows the normalization step, before application of non-linearity. See \cite{ioffe2015batch} for specifics. 

We performed experiments on MNIST, CIFAR-10, CIFAR-100, and SVHN datasets. We compared the performance of GBN transformations with 7 different deviation measures and statistics, including the conventional mean and standard deviation. As indicated in Table~\ref{risk_measure_table}, we utilized standard SD along with MAD, RSD, RBD, and SQD with $\alpha=.25, .5, \text{ and }.75$ which we denote by SQD1, SQD2, and SQD3 respectively. We omit WCD since centering w.r.t. $\sup x$ is obviously a poor choice when paired with ReLU. Subtracting $\sup x$ would make all points less than or equal to zero and the ReLU would send them all then to zero, producing an untrainable network without activations.

\subsection{MNIST}

\begin{figure}[!tb]
\centering
\begin{subfigure}[b]{0.48\columnwidth}
\centering
\includegraphics[width=\columnwidth]{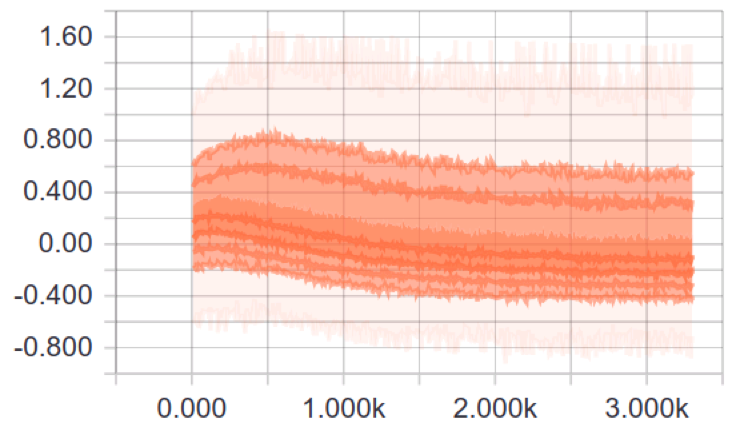}
\caption{\scriptsize{Before the standard BN}}
\end{subfigure}
\begin{subfigure}[b]{0.48\columnwidth}
\centering
\includegraphics[width=\columnwidth]{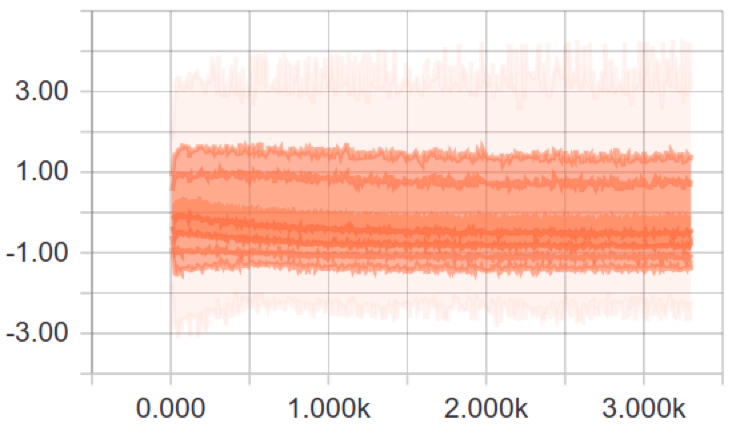}
\caption{\scriptsize{After the standard BN}}
\end{subfigure}
\begin{subfigure}[b]{0.48\columnwidth}
\centering
\includegraphics[width=\columnwidth]{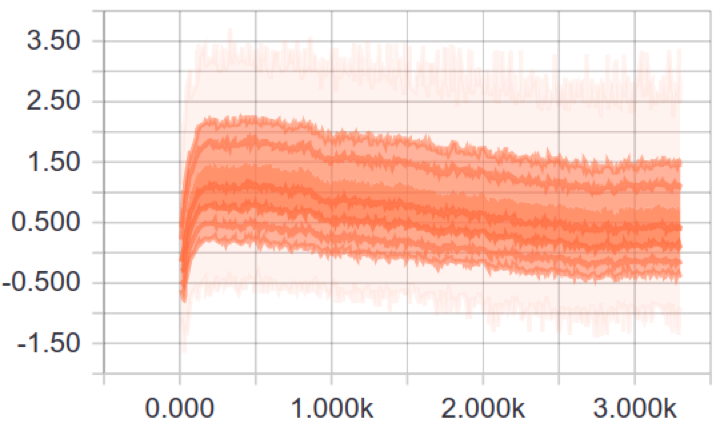}
\caption{\scriptsize{Before the GBN with SQD1}}
\end{subfigure}
\begin{subfigure}[b]{0.48\columnwidth}
\centering
\includegraphics[width=\columnwidth]{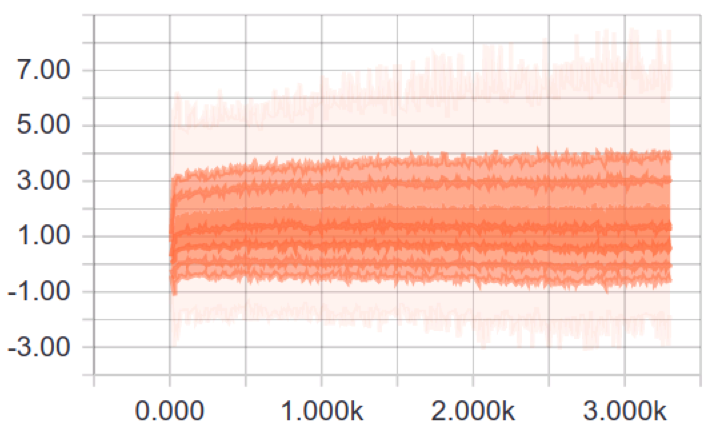}
\caption{\scriptsize{After the GBN with SQD1}}
\end{subfigure}
\begin{subfigure}[b]{0.48\columnwidth}
\centering
\includegraphics[width=\columnwidth]{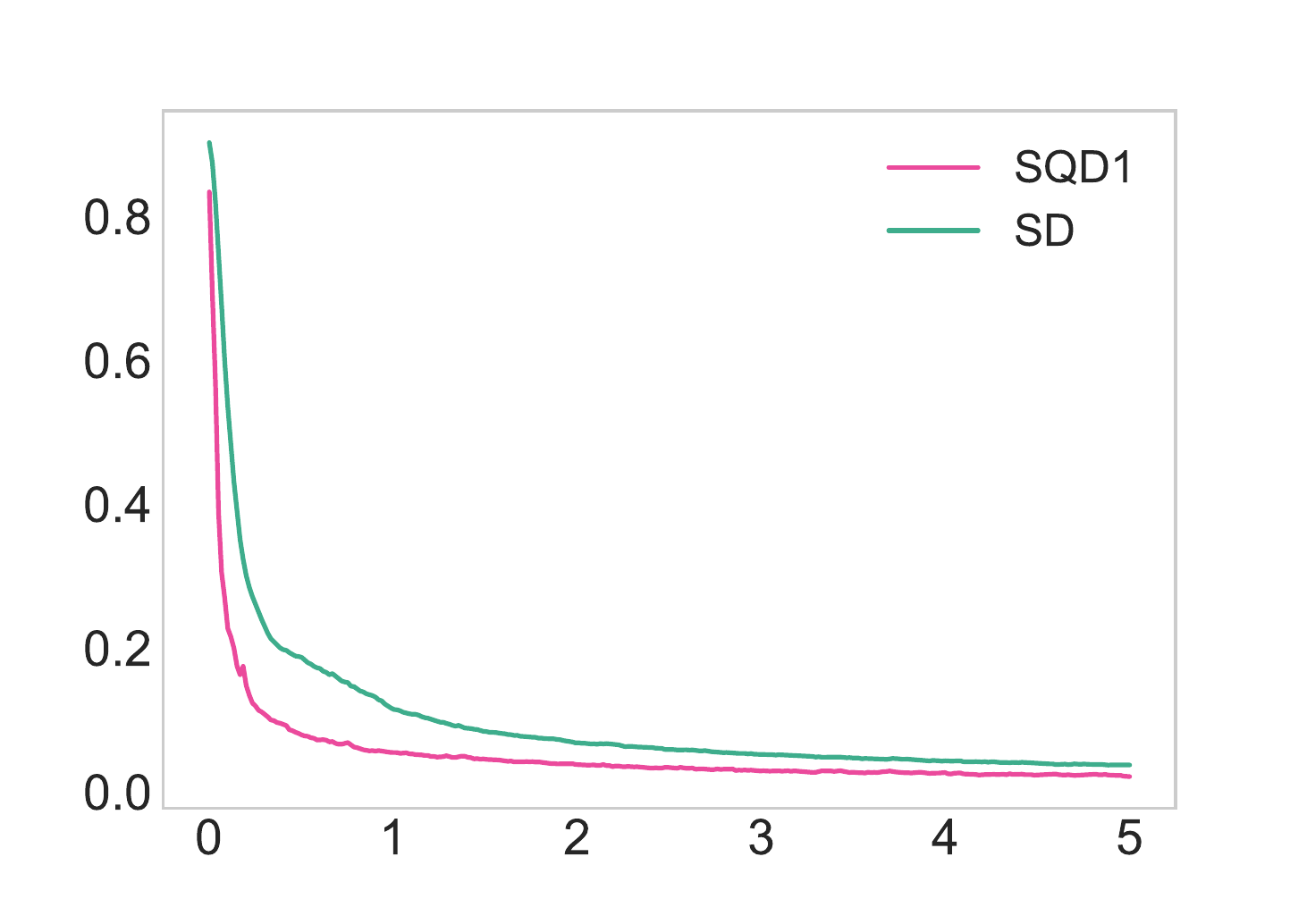}
\caption{\scriptsize{Error rate on held out test set}}
\end{subfigure}
\caption{ (a, b) The distribution evolution of standard BN on a selected feature along with the iteration steps. (c, d) The same figure over the GBD with SQD1 deviation measure. (e) The error rate of two settings on the held out test set. GBN with SQD1 help the network converges faster and also achieves better error rate that standard BN.}
\label{MNIST-Distribution}
\end{figure}

\subsubsection{GBN transformation over time}
To illustrate the effect that an asymmetric deviation measure and statistic have on the distribution of network activations when paired with ReLU, we observe the predictive error rate and the distribution over one feature before and after the GBN transformation. We conduct classification on MNIST \cite{lecun1998gradient} with neural network architecture LeNet with the input size of 28x28 and two convolutional layers with kernel size 5, and number of filters 20 and 50 respectively. The batch normalization is added after each of the convolutional layers and then followed by a ReLU non-linearity. The comparison is performed on standard BN and GBN with deviation measure SQD1, which has statistic equal to the $\alpha=.25$ quantile. We choose to observe one feature pixel of the second convolutional layer's feature map. 
Figure~\ref{MNIST-Distribution}(a,b) shows this feature's distribution density before and after standard BN. Figure~\ref{MNIST-Distribution}(c,d) shows the same feature's distribution density before and after applying GBN with SQD1. All the distributions before batch normalization exhibit significant change in terms of mean and variance. Both of the two normalization approaches removed the covariate shift effect and output a stabilized distribution over time. And after GBN with the deviation measure of SQD1, most of the values appear larger than 0 compared to the symmetric distribution of standard BN having the mean of 0. As one would expect, centering w.r.t. the $\alpha=.25$ quantile forces $\alpha\%$ of the activations to be less than zero before applying the non-linearity. In  Figure~\ref{MNIST-Distribution}(e), this consistent asymmetric distribution of the GBN's output helps it achieve faster convergence rate and better error rate compared to the standard BN.

\subsubsection{GBN performance on MNIST}
\begin{figure}[!tb]
\begin{center}
\begin{subfigure}[b]{0.48\columnwidth}
\centering
\includegraphics[width=\columnwidth]{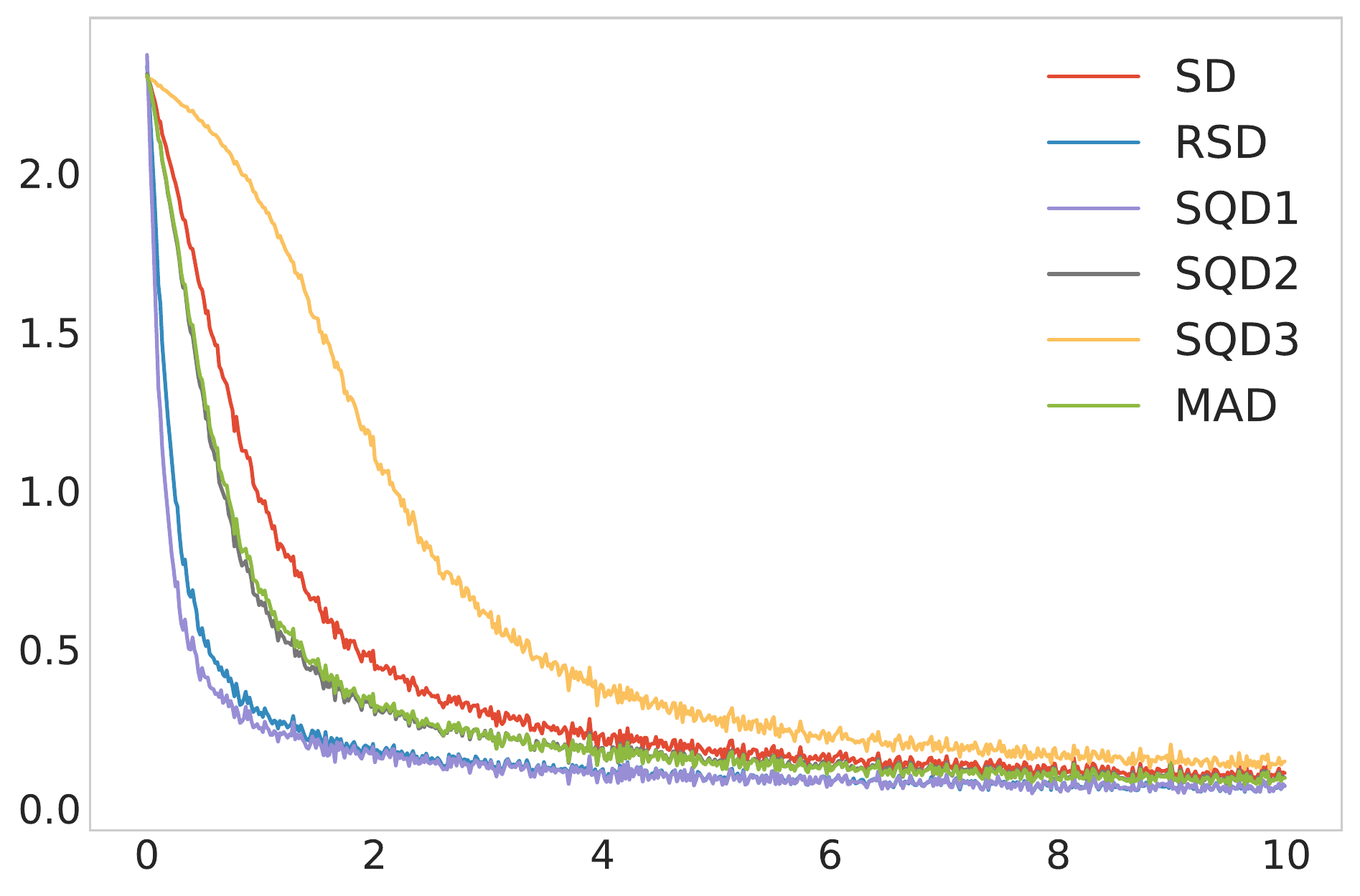}
\caption{\scriptsize{Training Loss}}
\end{subfigure}
\begin{subfigure}[b]{0.48\columnwidth}
\centering
\includegraphics[width=\columnwidth]{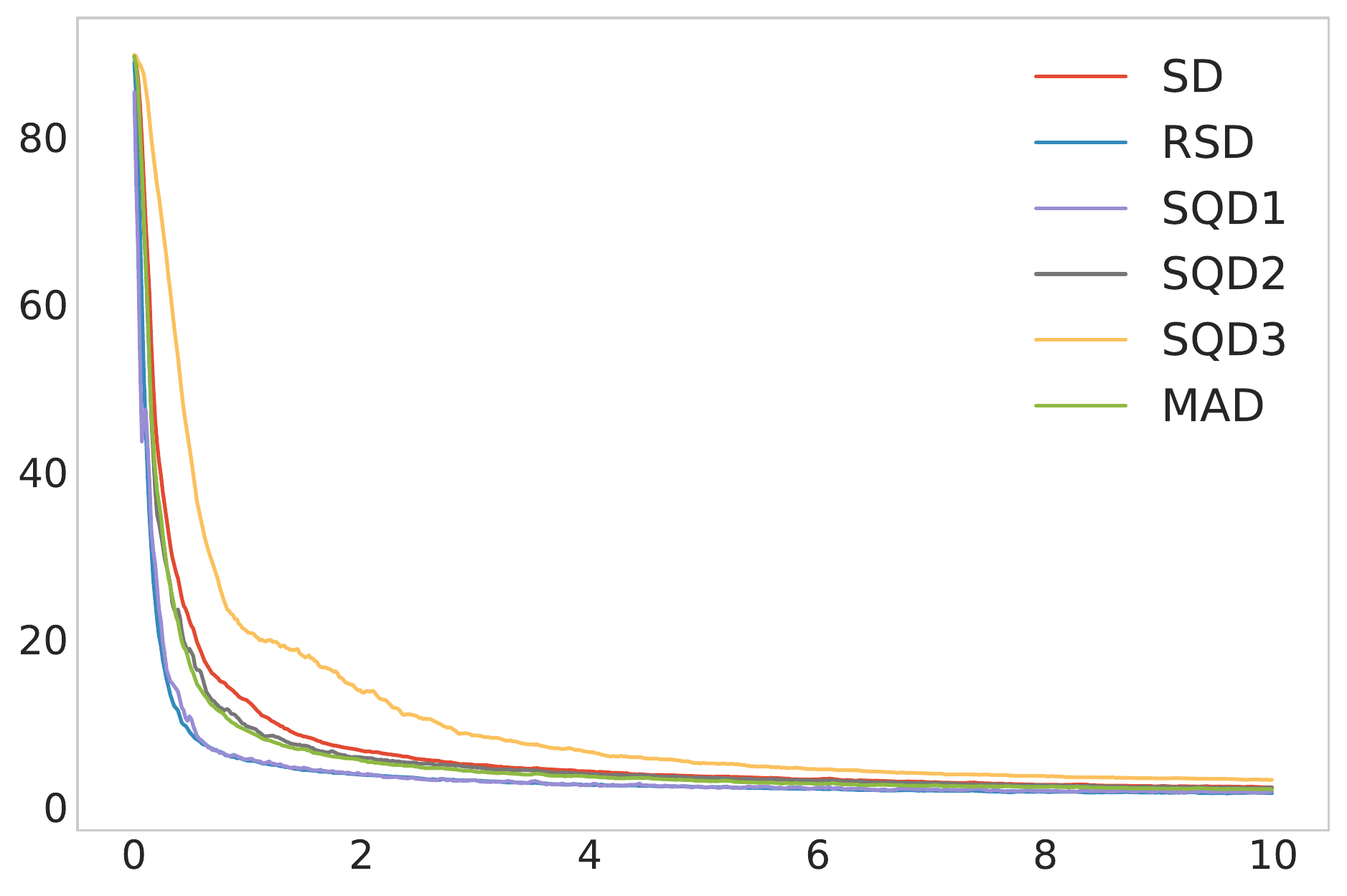}
\caption{\scriptsize{Test Error Rate(\%)}}
\end{subfigure}
\caption{Performance comparison of the MNIST classification with LeNet. X-axis: Number of epochs; Y-axis: Training loss/Test error rate.}
\label{MNIST-Performance}
\end{center}
\end{figure}

To compare the performance of various deviation measures and statistics on MNIST, we use the same experimental setting of neural network above with vanilla SGD as the optimizer, with learning rate equal to $.01$, and batch size equal to 1000. Figure~\ref{MNIST-Performance} shows the error rate of 6 different choices for deviation measure and statistic. All settings are evaluated on the training loss and test error rate. We see that GBN with SQD1, RSD, SQD2, and MAD all perform better than standard BN in terms of converge rate and test error rate. And GBN with deviation measures of SQD1 and RSD converge remarkably faster than others. 

\subsection{CIFAR-10, CIFAR-100, and SVHN}
We compare the performance and convergence rate on the CIFAR-10, CIFAR-100, and Street View House Numbers (SVHN) datasets. The CIFAR-10 and CIFAR-100 dataset consist of 60,000 tiny color images (32x32) with 10 and 100 classes respectively for image recognition task~\cite{krizhevsky2009learning}. The SVHN Dataset consists of Google Street View images with 10 house digit classes~\cite{netzer2011reading}. 

\begin{figure}[!tb]
\begin{center}
\begin{subfigure}[b]{0.48\columnwidth}
\centering
\includegraphics[width=\columnwidth]{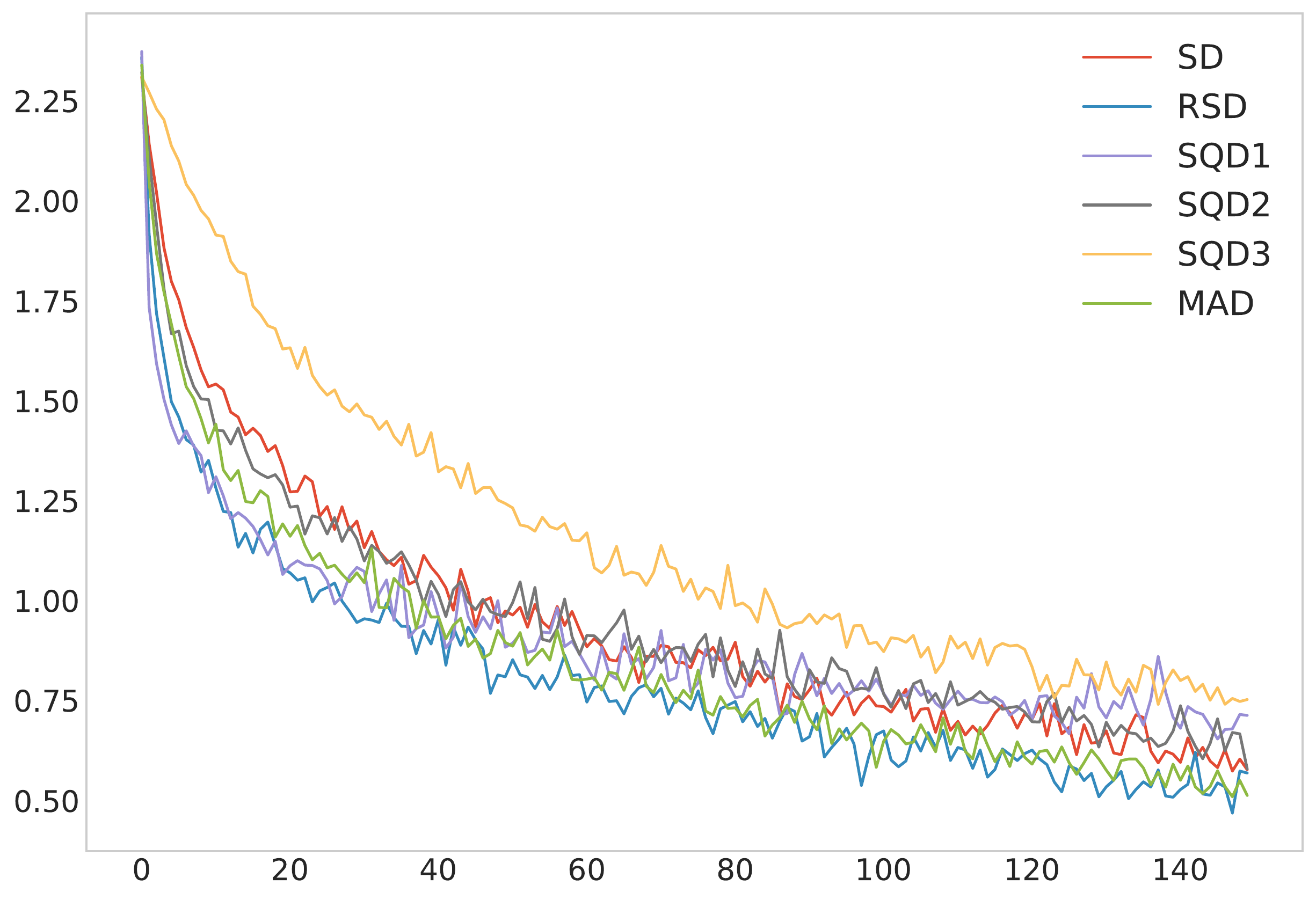}
\caption{\scriptsize{Training Loss}}
\end{subfigure}
\begin{subfigure}[b]{0.48\columnwidth}
\centering
\includegraphics[width=\columnwidth]{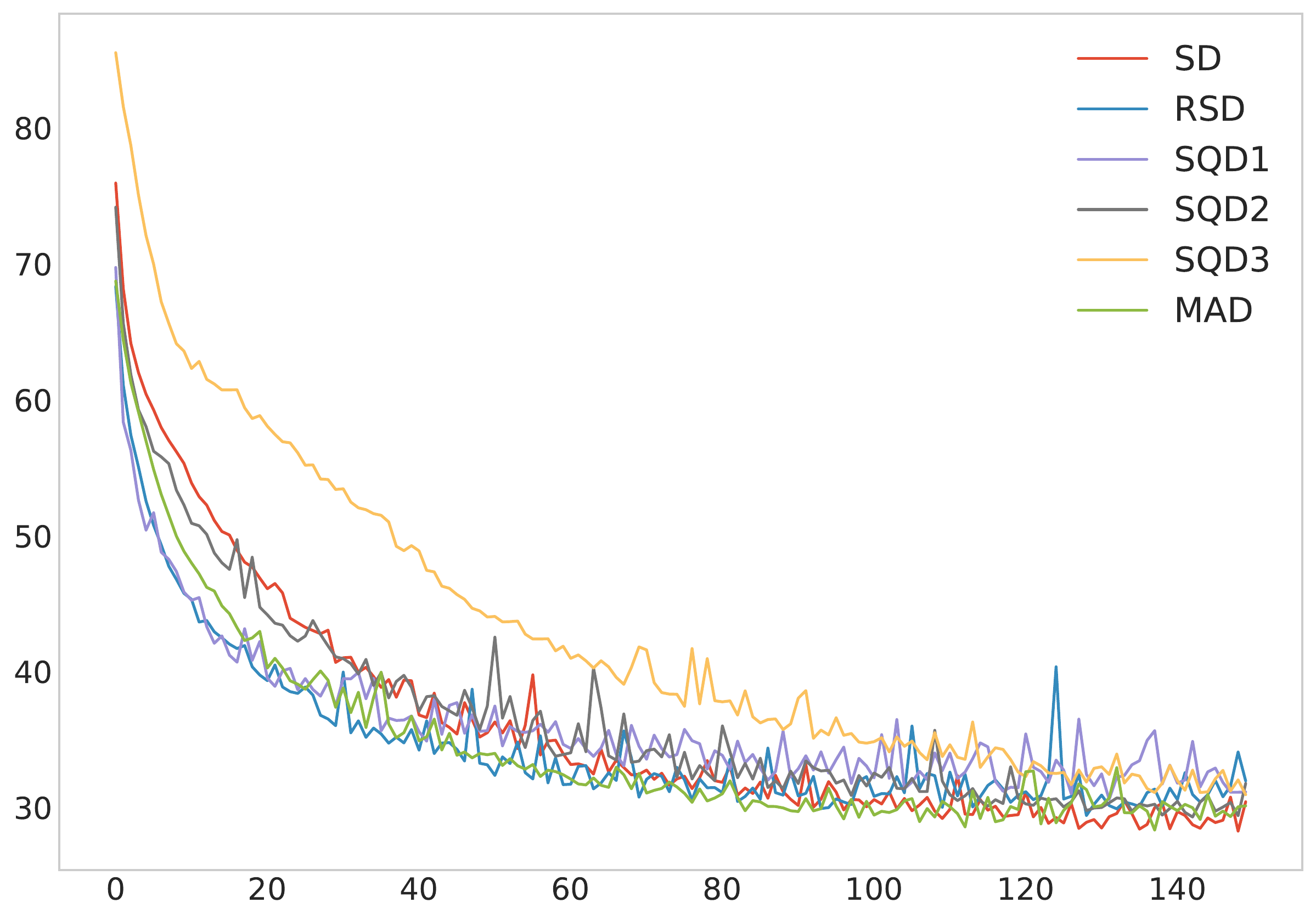}
\caption{\scriptsize{Test Error Rate}}
\end{subfigure}
\caption{Performance comparison on the CIFAR-10 dataset using LeNet. X-axis: Number of epochs; Y-axis: Training loss/Test error rate (\%).  We use learning rate 0.01 and batch size 1024.}
\label{fig:cifar10}
\end{center}
\end{figure}

\begin{figure}[!tb]
\begin{center}
\begin{subfigure}[b]{0.48\columnwidth}
\centering
\includegraphics[width=\columnwidth]{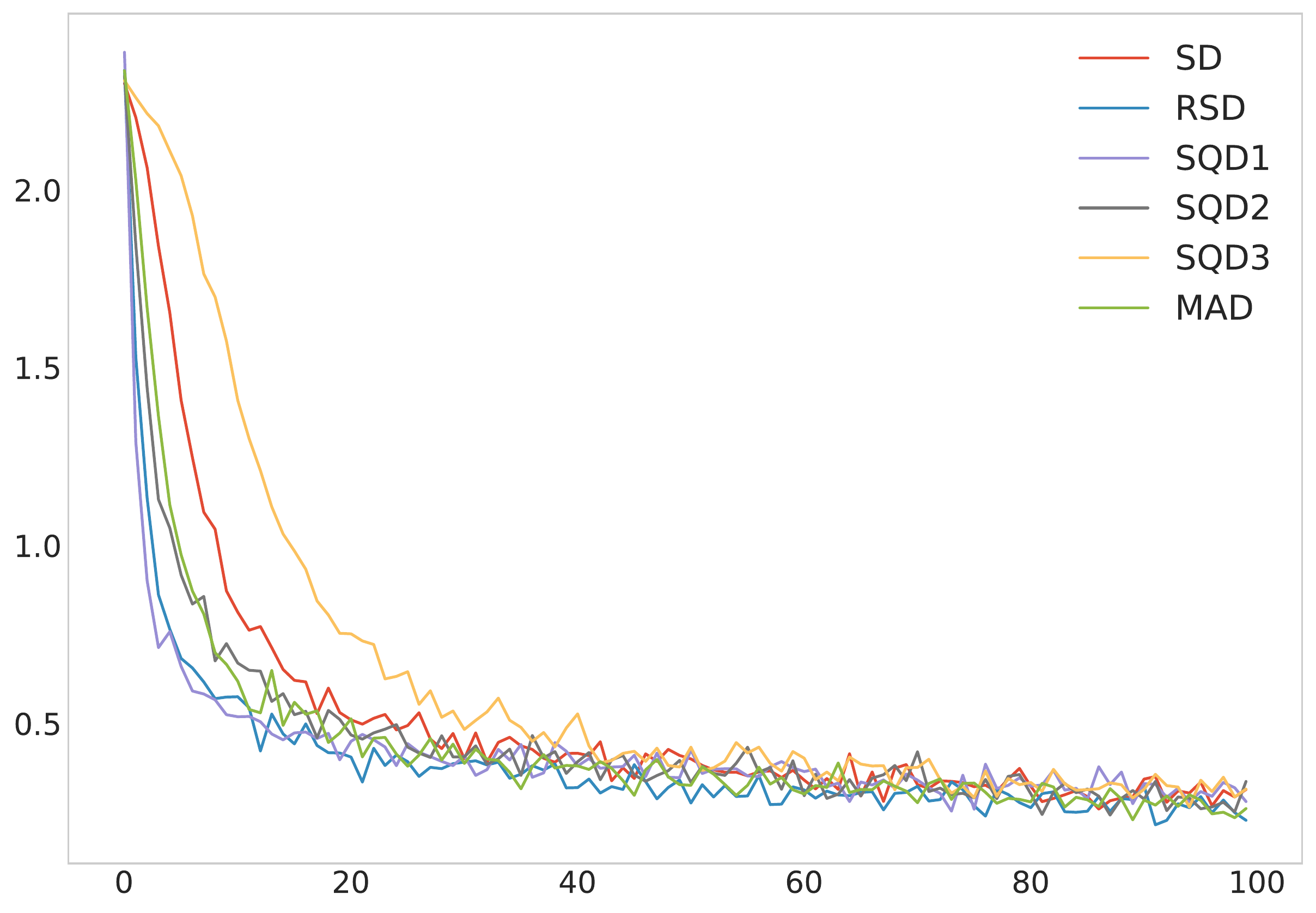}
\caption{\scriptsize{Training Loss}}
\end{subfigure}
\begin{subfigure}[b]{0.48\columnwidth}
\centering
\includegraphics[width=\columnwidth]{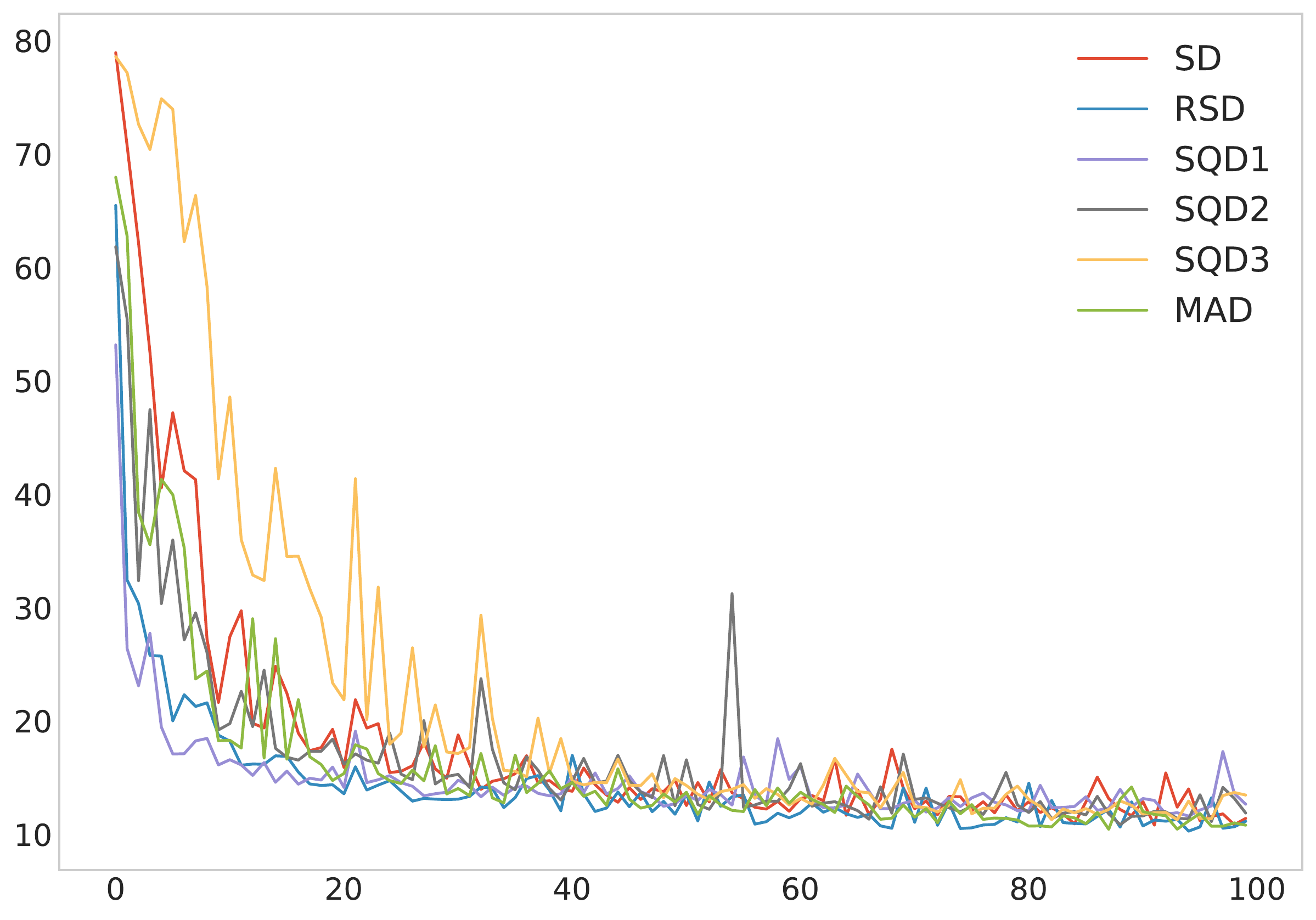}
\caption{\scriptsize{Test Error Rate}}
\end{subfigure}
\caption{Performance comparison on the SVHN dataset. X-axis: Number of epochs; Y-axis: Training loss/Test error rate (\%). We use learning rate 0.01 and batch size 1024.}
\label{fig:svhn}
\end{center}
\end{figure}

We trained LeNet networks on the CIFAR-10 (200 epochs) and the SVHN (160 epochs) datasets. The setting is set similar to that used with MNIST dataset: SGD with learning rate 0.1 and 0.01, batch size 1024.
Figure~\ref{fig:cifar10} and Figure~\ref{fig:svhn} illustrate the performance comparison of six different choices of deviation measure on the CIFAR-10 and SVHN datasets respectively. 

\begin{figure}[!tb]
\begin{center}
\begin{subfigure}[b]{0.48\columnwidth}
\centering
\includegraphics[width=\columnwidth]{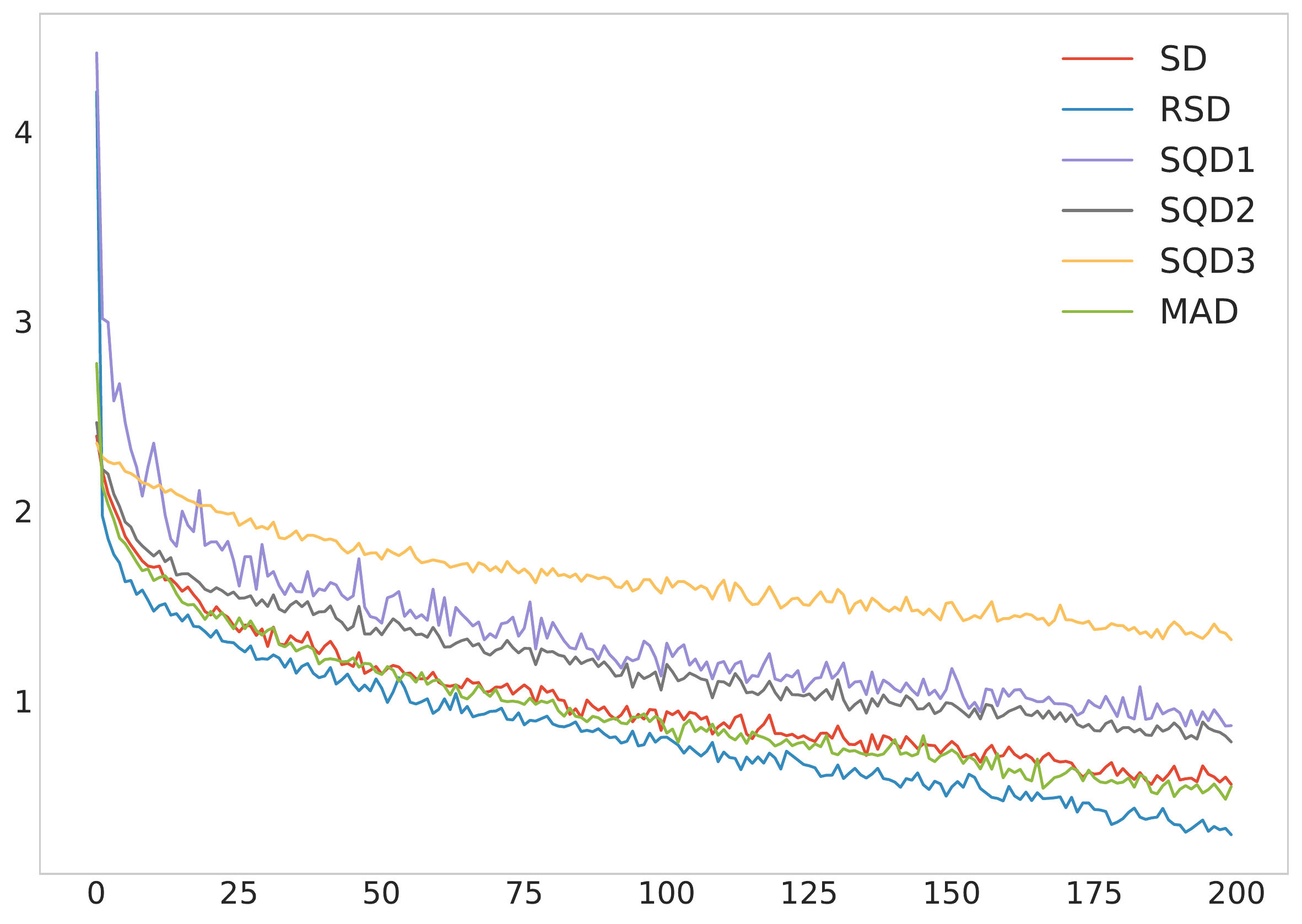}
\caption{\scriptsize{Training Loss}}
\end{subfigure}
\begin{subfigure}[b]{0.48\columnwidth}
\centering
\includegraphics[width=\columnwidth]{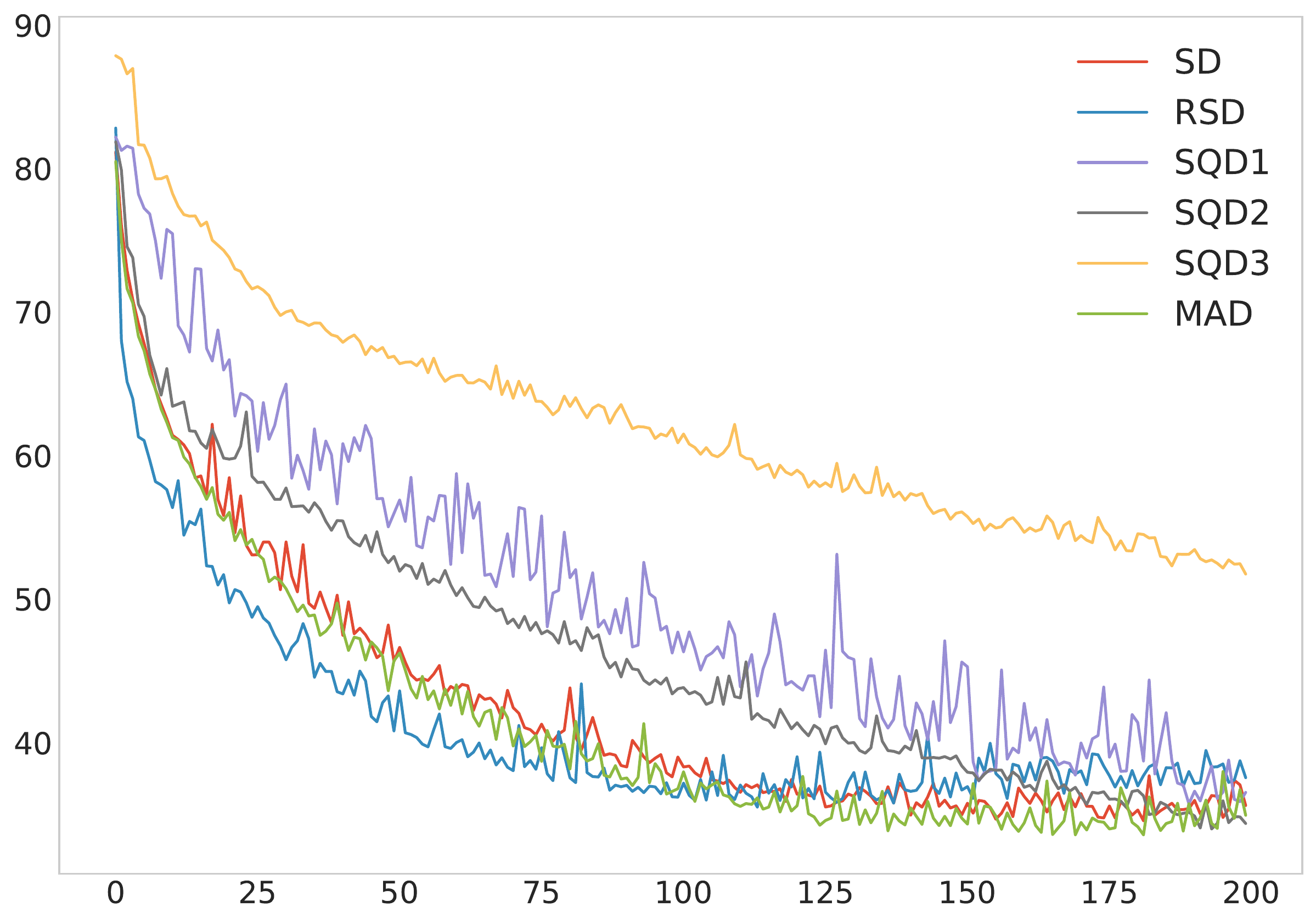}
\caption{\scriptsize{Test Error Rate}}
\end{subfigure}
\caption{Performance comparison on the CIFAR-10 dataset using ResNet20. X-axis: Number of epochs; Y-axis: Training loss/Test error rate (\%). We use learning rate 0.01 and batch size 1024.}
\label{fig:cifar10-resnet}
\end{center}
\end{figure}

\begin{figure}[!tb]
\begin{center}
\begin{subfigure}[b]{0.48\columnwidth}
\centering
\includegraphics[width=\columnwidth]{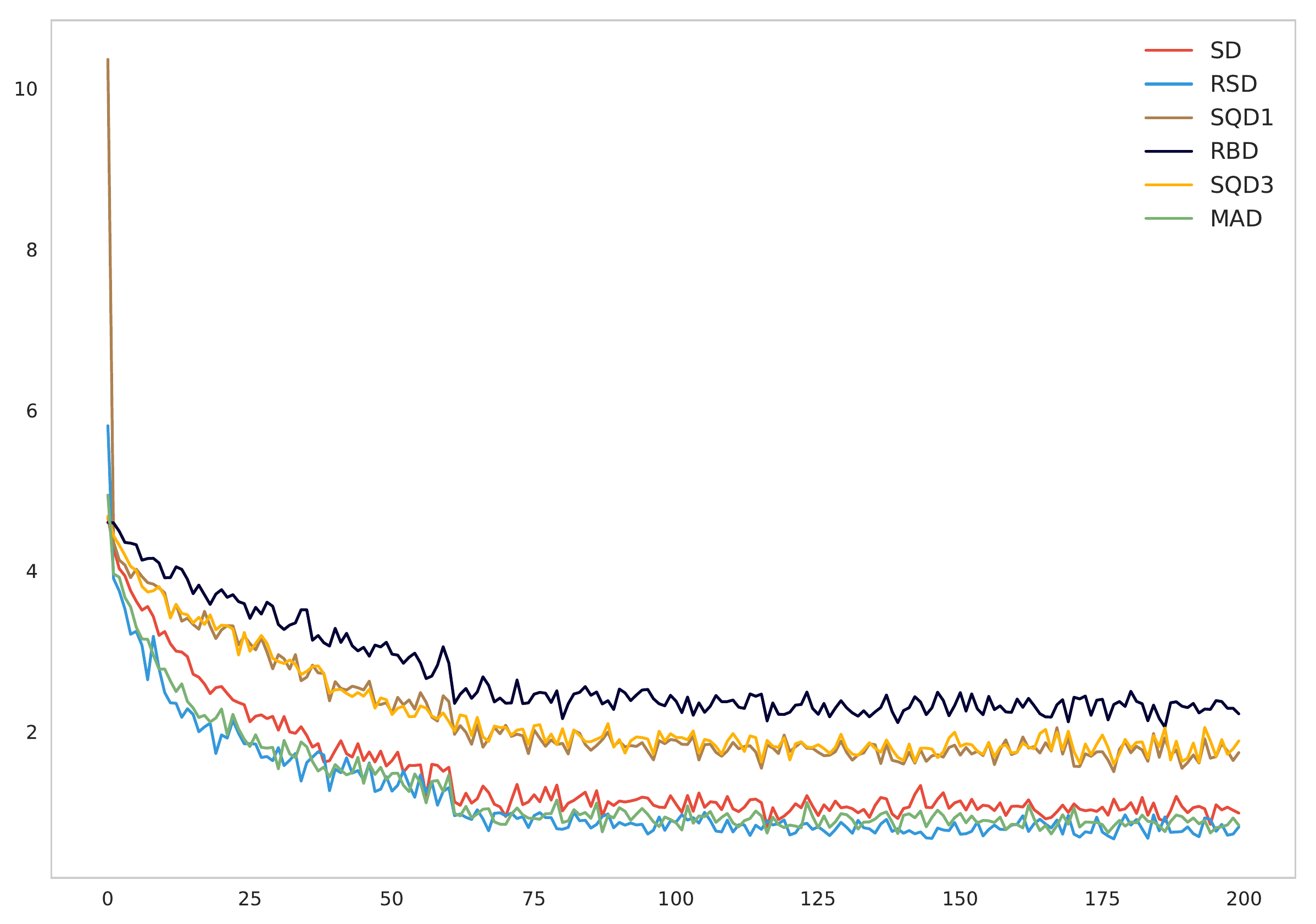}
\caption{\scriptsize{Training Loss}}
\end{subfigure}
\begin{subfigure}[b]{0.48\columnwidth}
\centering
\includegraphics[width=\columnwidth]{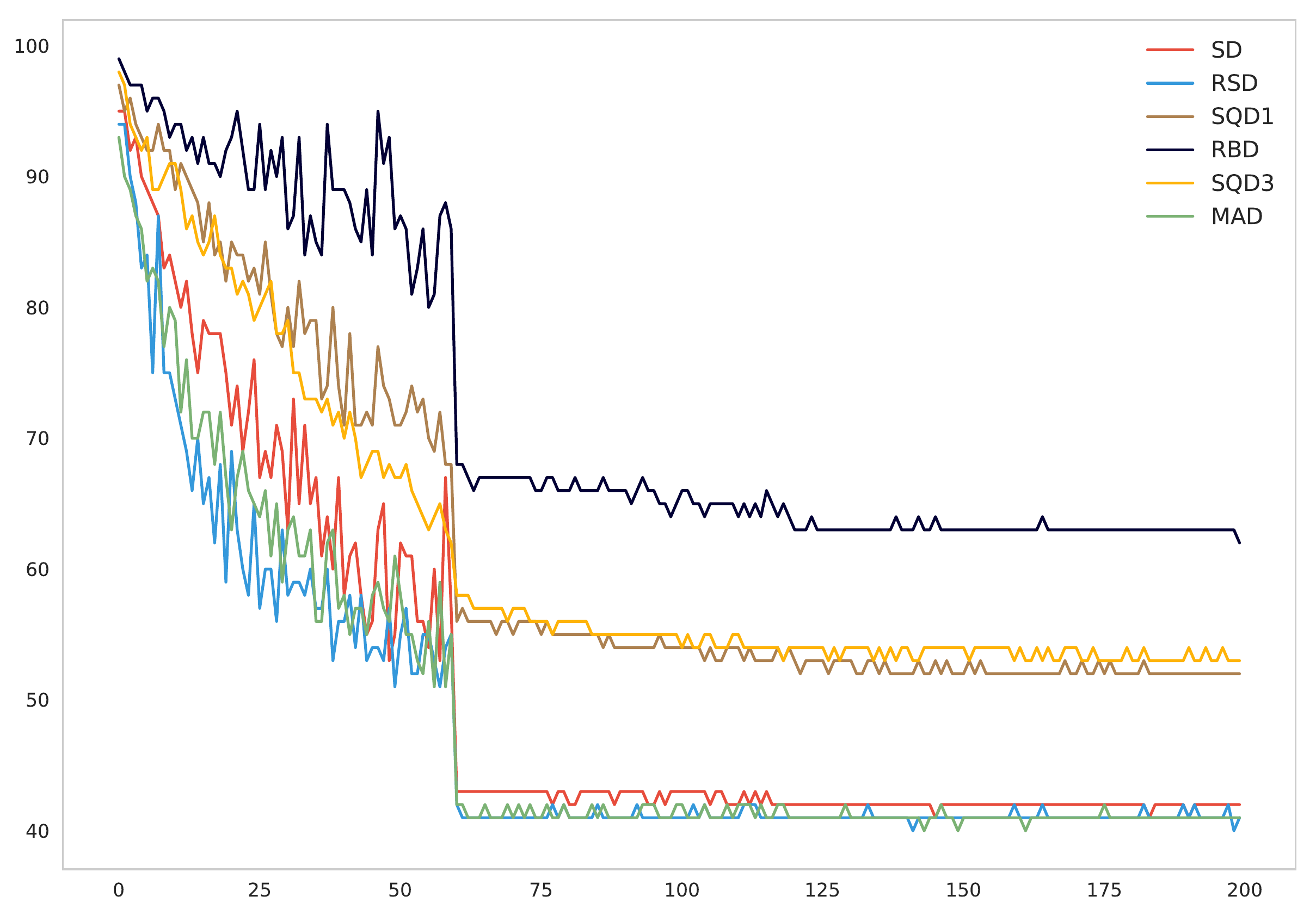}
\caption{\scriptsize{Test Error Rate}}
\end{subfigure}
\caption{Performance comparison on the CIFAR-100 dataset using ResNet20. X-axis: Number of epochs; Y-axis: Training loss/Test error rate (\%). We use learning rate 0.1 and batch size 256.}
\label{fig:cifar100-resnet}
\end{center}
\end{figure}

We also train a ResNet architecture with 20 layers (exactly the same architecture and settings used in~\cite{he2015delving}) for 200 epochs on the CIFAR-10 and  CIFAR-100 dataset.
We trained the ResNet with and without data augmentation (i.e., random crop and random horizontal flip). For the CIFAR-10 dataset, we observe that with data augmentation, the proposed methods achieve performance similar to standard BN. However, if we do not augment data (less symmetric distribution), both RSD and MAD perform better than standard BN (Figure~\ref{fig:cifar10-resnet}). For the CIFAR-100 dataset, even with data augmentation, RSD and MAD outperform standard BN (Figure~\ref{fig:cifar100-resnet}).

\begin{table*}[!tb]
\centering
\scriptsize
\begin{tabular}{|c|c|c|c|c|c|c|c|c|c|c|}
\hline
Dataset   & Architecture & LR & BS & SD            & MAD           & RSD           & RBD           & SQD1          & SQD2          & SQD3          \\ \hline
CIFAR-10  & LeNet        & 0.1           & 256        & 29.92 ($\pm$0.71) & 29.72 ($\pm$0.71) & 30.34 ($\pm$0.34) & 29.17 ($\pm$0.47) & 32.44 ($\pm$0.90) & 28.56 ($\pm$0.58) & \textbf{28.37} ($\pm$0.24) \\ \hline
CIFAR-10  & LeNet        & 0.1           & 1024       & 28.74 ($\pm$0.36) & 29.03 ($\pm$0.37) & 29.19 ($\pm$0.45) & 31.16 ($\pm$0.30) & 46.64 ($\pm$4.05) & 27.23 ($\pm$0.19) & \textbf{27.17} ($\pm$0.27) \\ \hline
CIFAR-10  & LeNet        & 0.1           & 2048       & 27.35 ($\pm$0.15) & 27.75 ($\pm$0.29) & 28.14 ($\pm$0.13) & 36.76 ($\pm$0.77) & 48.65 ($\pm$6.27) & 27.24 ($\pm$0.35) & \textbf{26.19} ($\pm$0.24) \\ \hline
CIFAR-10  & LeNet        & 0.01          & 256        & 28.28 ($\pm$0.41) & 28.97 ($\pm$0.38) & 30.45 ($\pm$0.33) & 33.29 ($\pm$0.32) & 28.49 ($\pm$0.31) & 27.33 ($\pm$0.23) & \textbf{26.54} ($\pm$0.29) \\ \hline
CIFAR-10  & ResNet20     & 0.1           & 256        & 22.99 ($\pm$0.64) & 23.39 ($\pm$0.17) & 23.11 ($\pm$0.42) & \textbf{15.75} ($\pm$0.15) & 18.05 ($\pm$0.51) & 20.49 ($\pm$0.57) & 19.82 ($\pm$0.35) \\ \hline
CIFAR-10  & ResNet20     & 0.1           & 1024       & 29.69 ($\pm$0.45) & 29.9 ($\pm$0.56)  & 29.19 ($\pm$0.66) & \textbf{23.08} ($\pm$0.52) & 30.96 ($\pm$2.01) & 24.32 ($\pm$0.97) & 26.51 ($\pm$0.49) \\ \hline
CIFAR-10  & ResNet20     & 0.1           & 2048       & 30.99 ($\pm$0.78) & \textbf{30.46} ($\pm$0.54) & 30.48 ($\pm$0.85) & 32.46 ($\pm$0.41) & 41.35 ($\pm$1.19) & 30.47 ($\pm$0.66) & 36.57 ($\pm$0.32) \\ \hline
CIFAR-10  & ResNet20     & 0.01          & 256        & 33.33 ($\pm$1.05) & 33.12 ($\pm$0.25) & 34.59 ($\pm$0.78) & 32.39 ($\pm$0.55) & \textbf{25.58} ($\pm$0.31) & 26.84 ($\pm$0.48) & 37.48 ($\pm$0.85) \\ \hline
CIFAR-100 & ResNet20     & 0.1           & 256        & 58.38 ($\pm$0.28) & 58.41 ($\pm$0.43) & 57.81 ($\pm$0.34) & 50.38 ($\pm$0.37) & \textbf{46.50} ($\pm$0.95) & 47.98 ($\pm$0.16) & 49.32 ($\pm$0.53) \\ \hline
CIFAR-100 & ResNet20     & 0.1           & 1024       & 60.26 ($\pm$0.57) & 61.66 ($\pm$0.70) & 63.96 ($\pm$1.18) & 77.96 ($\pm$0.25) & \textbf{54.55} ($\pm$0.86) & 54.80 ($\pm$0.98) & 64.61 ($\pm$0.60) \\ \hline
CIFAR-100 & ResNet20     & 0.1           & 2048       & 65.91 ($\pm$0.69) & \textbf{61.06} ($\pm$0.88) & 62.20 ($\pm$0.94) & 87.8 ($\pm$0.77)  & 64.08 ($\pm$1.23) & 63.3 ($\pm$0.64)  & 76.50 ($\pm$0.67) \\ \hline
CIFAR-100 & ResNet20     & 0.01          & 256        & 69.09 ($\pm$0.63) & 64.15 ($\pm$0.57) & 65.39 ($\pm$0.68) & 88.15 ($\pm$0.51) & \textbf{55.75} ($\pm$0.45) & 64.50 ($\pm$0.60) & 78.83 ($\pm$0.80) \\ \hline
SVHN      & LeNet        & 0.1           & 256        & 10.49 ($\pm$0.17) & 10.61 ($\pm$0.22) & 10.80 ($\pm$0.18) & 11.05 ($\pm$0.22) & 15.10 ($\pm$0.83) & 10.98 ($\pm$0.20) & \textbf{10.49} ($\pm$0.22) \\ \hline
SVHN      & LeNet        & 0.1           & 1024       & 9.918 ($\pm$0.17) & 10.09 ($\pm$0.08) & 10.29 ($\pm$0.10) & 11.53 ($\pm$0.21) & 14.99 ($\pm$0.59) & 9.968 ($\pm$0.13) & \textbf{9.447} ($\pm$0.16) \\ \hline
SVHN      & LeNet        & 0.1           & 2048       & 9.559 ($\pm$0.14) & 9.681 ($\pm$0.17) & 10.17 ($\pm$0.17) & 13.26 ($\pm$0.28) & 19.27 ($\pm$3.93) & 9.593 ($\pm$0.19) & \textbf{9.413} ($\pm$0.32) \\ \hline
SVHN      & LeNet        & 0.01          & 256        & 9.023 ($\pm$0.21) & 9.323 ($\pm$0.16) & 10.07 ($\pm$0.24) & 13.07 ($\pm$0.36) & 10.37 ($\pm$0.17) & 9.248 ($\pm$0.14) & \textbf{8.922} ($\pm$0.09) \\ \hline
\end{tabular}
\caption{Performance Comparison of Test Error Rate (\%) without Data Augmentation.}
\label{tab:cifar10-svhn}
\end{table*}

\begin{table*}[!tb]
\centering
\scriptsize
\begin{tabular}{|c|c|c|c|c|c|c|c|c|c|c|}
\hline
Dataset   & Architecture & LR & BS & SD            & MAD           & RSD           & RBD           & SQD1          & SQD2          & SQD3          \\ \hline
CIFAR-10  & LeNet        & 0.1           & 256        & 22.12 ($\pm$0.10) & \textbf{21.96} ($\pm$0.15) & 22.24 ($\pm$0.12) & 30.23 ($\pm$0.33) & 38.55 ($\pm$1.98) & 24.91 ($\pm$0.27) & 24.57 ($\pm$0.19) \\ \hline
CIFAR-10  & LeNet        & 0.1           & 1024       & 23.92 ($\pm$0.16) & 23.79 ($\pm$0.16) & \textbf{23.76} ($\pm$0.16) & 37.14 ($\pm$0.27) & 50.97 ($\pm$8.63) & 27.4 ($\pm$0.31)  & 26.70 ($\pm$0.20) \\ \hline
CIFAR-10  & LeNet        & 0.1           & 2048       & 25.57 ($\pm$0.25) & 25.54 ($\pm$0.24) & \textbf{25.25} ($\pm$0.11) & 44.15 ($\pm$0.79) & 48.29 ($\pm$6.84) & 30.11 ($\pm$0.40) & 28.55 ($\pm$0.37) \\ \hline
CIFAR-10  & LeNet        & 0.01          & 256        & 25.49 ($\pm$0.17) & 25.11 ($\pm$0.25) & \textbf{24.52} ($\pm$0.17) & 40.24 ($\pm$0.48) & 28.46 ($\pm$0.12) & 26.89 ($\pm$0.12) & 27.41 ($\pm$0.41) \\ \hline
CIFAR-10  & ResNet20     & 0.1           & 256        & \textbf{11.69} ($\pm$0.19) & 11.76 ($\pm$0.19) & 11.84 ($\pm$0.27) & 13.57 ($\pm$0.19) & 17.09 ($\pm$1.03) & 15.37 ($\pm$0.29) & 14.80 ($\pm$0.21) \\ \hline
CIFAR-10  & ResNet20     & 0.1           & 1024       & 15.33 ($\pm$0.21) & \textbf{15.20} ($\pm$0.15) & 15.77 ($\pm$0.41) & 26.2 ($\pm$0.13)  & 32.24 ($\pm$0.97) & 23.77 ($\pm$1.00) & 24.78 ($\pm$0.99) \\ \hline
CIFAR-10  & ResNet20     & 0.1           & 2048       & 20.09 ($\pm$0.35) & \textbf{19.41} ($\pm$0.32) & 19.43 ($\pm$0.57) & 38.35 ($\pm$1.12) & 44.83 ($\pm$1.78) & 31.25 ($\pm$1.27) & 36.51 ($\pm$0.76) \\ \hline
CIFAR-10  & ResNet20     & 0.01          & 256        & 21.30 ($\pm$0.67) & 21.58 ($\pm$0.23) & \textbf{20.24} ($\pm$0.41) & 35.26 ($\pm$0.97) & 29.04 ($\pm$0.55) & 26.73 ($\pm$0.75) & 39.28 ($\pm$1.08) \\ \hline
CIFAR-100 & ResNet20     & 0.1           & 256        & 38.47 ($\pm$0.08) & 38.41 ($\pm$0.29) & \textbf{38.15} ($\pm$0.26) & 53.03 ($\pm$0.46) & 45.99 ($\pm$0.45) & 44.86 ($\pm$0.43) & 46.61 ($\pm$0.71) \\ \hline
CIFAR-100 & ResNet20     & 0.1           & 1024       & 52.64 ($\pm$0.55) & 46.71 ($\pm$0.15) & \textbf{44.58} ($\pm$0.30) & 80.53 ($\pm$0.83) & 58.05 ($\pm$0.39) & 56.19 ($\pm$0.41) & 65.96 ($\pm$0.67) \\ \hline
CIFAR-100 & ResNet20     & 0.1           & 2048       & 65.57 ($\pm$0.27) & 56.70 ($\pm$0.32) & \textbf{52.08} ($\pm$0.50) & 88.98 ($\pm$0.71) & 68.03 ($\pm$0.91) & 65.75 ($\pm$0.41) & 77.27 ($\pm$0.21) \\ \hline
CIFAR-100 & ResNet20     & 0.01          & 256        & 68.75 ($\pm$0.32) & 59.65 ($\pm$0.43) & \textbf{54.40} ($\pm$0.25) & 89.19 ($\pm$0.46) & 58.53 ($\pm$0.36) & 67 ($\pm$0.40)    & 79.84 ($\pm$0.45) \\ \hline
SVHN      & LeNet        & 0.1           & 256        & \textbf{12.09} ($\pm$0.15) & 12.10 ($\pm$0.23) & 12.11 ($\pm$0.33) & 22.54 ($\pm$0.39) & 29.95 ($\pm$4.83) & 14.11 ($\pm$0.27) & 13.59 ($\pm$0.11) \\ \hline
SVHN      & LeNet        & 0.1           & 1024       & \textbf{13.84} ($\pm$0.25) & 13.84 ($\pm$0.14) & 13.86 ($\pm$0.25) & 28.12 ($\pm$0.44) & 36.41 ($\pm$3.86) & 16.69 ($\pm$0.20) & 15.00 ($\pm$0.29) \\ \hline
SVHN      & LeNet        & 0.1           & 2048       & 15.85 ($\pm$0.58) & 15.57 ($\pm$0.47) & \textbf{15.33} ($\pm$0.31) & 36.87 ($\pm$1.27) & 50.60 ($\pm$7.54) & 19.28 ($\pm$0.48) & 17.18 ($\pm$0.77) \\ \hline
SVHN      & LeNet        & 0.01          & 256        & 15.01 ($\pm$0.44) & 14.89 ($\pm$0.35) & \textbf{14.39} ($\pm$0.27) & 32.96 ($\pm$0.85) & 18.66 ($\pm$0.47) & 16.13 ($\pm$0.40) & 15.72 ($\pm$0.30) \\ \hline
\end{tabular}
\caption{Performance Comparison of Test Error Rate (\%) with Data Augmentation.}
\label{tab:cifar10-svhn2}
\end{table*}


Most methods (MAD, SQD1, SQD2, and RSD) converge faster than standard SD. The error rates of these methods are similar, but with slight improvement achieved by the proposed alternative deviation measures. Table~\ref{tab:cifar10-svhn} and \ref{tab:cifar10-svhn2} contain more detailed results regarding the error rate achieved for a few extra settings. LR denotes learning rate, BS denotes batch size.
We run every setting five times, each time using a different shuffle of the training data, and report the mean and the standard deviation of the best test error rate achieved during training. We also run these settings with and without data augmentation.
Faster converging alternatives achieve similar error rates. In addition, we see that these alternative deviation measures can often lead to increased accuracy when compared to SD, especially when data augmentation is not used. 

\subsection{Discussion}
When choosing $\mathcal{S}$ and $\mathcal{D}$, it is important to consider their estimation properties. For example, it is well-known that empirical estimates of the mean are more stable, and converge more quickly to the true mean, than empirical estimates of the superquantile. This also applies to SD and one-sided deviation measures like RSD. Clearly, since only one side of the distribution is involved, more samples will be needed for accurate, low variance estimation of asymmetric (one-sided) deviation measures or statistics.
Compared with small batch size, we observe that training with large batch size improves the convergence rate. However, this small-batch degradation is not a new consideration and has been observed with standard BN. \cite{ioffe2017batch} discusses this issue and shows that there do exist techniques to help alleviate this affect for BN. Although we leave this discussion to future work, it would seem straightforward to apply the same techniques to GBN in general.

\section{Conclusion}
In this paper, we have proposed a generalized variant of batch normalization which can be used to improve the convergence rate and, often, the error rate compared to vanilla batch normalization. As a generalization, we show that there are many other natural choices for the scaling and centering factors which we pose as general deviation measures and statistics. We also show that conventional normalization is not optimal if followed by the ReLU non-linearity and we provide alternatives that are justified both intuitively and theoretically, showing also that these new methods increase convergence speed experimentally.

\section{Acknowledgement}
The authors would like to thank anonymous reviewers whose suggestions help improve the quality of the paper. 
This research was supported in part by National Science Foundation (CNS-1624782, CNS-1747783), National Institutes of Health (R01-GM110240), and Industrial Members of NSF Center for Big Learning (CBL).

\bibliographystyle{aaai}
\bibliography{buffered_relu.bib,deep.bib}

\end{document}